\documentclass[journal]{IEEEtran}

\usepackage{amsfonts}
\usepackage{setspace}

\usepackage{setspace}
\usepackage{amssymb}
\usepackage{epsfig}
\usepackage{graphicx}
\usepackage{xcolor}
\usepackage{algorithm}
\usepackage{algorithmicx}
\usepackage{algpseudocode}
\usepackage{amsmath}
\usepackage{cite}

\floatname{algorithm}{Algorithm}

\usepackage[ps2pdf,bookmarks=false,
bookmarksnumbered=false, 
bookmarksopen=false,
colorlinks=false]{}

\usepackage[utf8]{inputenc}
\usepackage[english]{babel}
\usepackage{amsthm}
\theoremstyle{definition}

\usepackage{tikz}
\newsavebox{\tempbox}

\usepackage{multicol}

\IEEEoverridecommandlockouts\IEEEpubid{\makebox[\columnwidth]{ 978-1-6654-3540-6/22~\copyright~2022 IEEE \hfill} \hspace{\columnsep}\makebox[\columnwidth]{ }}

\begin{document}

\title{Communication-Efficient and Privacy-Preserving Feature-based Federated Transfer Learning}
\author{{Feng Wang,  M. Cenk Gursoy and Senem Velipasalar}
\thanks{The authors are with the Department of Electrical
		Engineering and Computer Science, Syracuse University, Syracuse, NY, 13244. E-mail: fwang26@syr.edu, mcgursoy@syr.edu, svelipas@syr.edu}}
\maketitle

\thispagestyle{empty}

\begin{abstract}
Federated learning has attracted growing interest as it preserves the clients' privacy. As a variant of federated learning, federated transfer learning utilizes the knowledge from similar tasks and thus has also been intensively studied. However, due to the limited radio spectrum, the communication efficiency of federated learning via wireless links is critical since some tasks may require thousands of Terabytes of uplink payload. In order to improve the communication efficiency, we in this paper propose the feature-based federated transfer learning as an innovative approach to reduce the uplink payload by more than five orders of magnitude compared to that of existing approaches. We first introduce the system design in which the extracted features and outputs are uploaded instead of parameter updates, and then determine the required payload with this approach and provide comparisons with the existing approaches. Subsequently, we analyze the random shuffling scheme that preserves the clients' privacy. Finally, we evaluate the performance of the proposed learning scheme via experiments on an image classification task to show its effectiveness. \footnote{Code implementation available at: https://github.com/wfwf10/Feature-based-Federated-Transfer-Learning.}
\end{abstract}


\section{Introduction}
Federated learning (FL), as a type of distributed learning, preserves data privacy by exchanging model parameter updates and keeping the datasets local \cite{konevcny2016federated}. Typically, FL considers a central parameter server (PS) that orchestrates many clients (i.e., participating devices, up to $10^{10}$ of them) to collaboratively train a deep neural network (DNN) model \cite{kairouz2019advances} by aggregating the updates on the weights and biases from each client. The widespread use of mobile phones and tablets with sufficient computational power and wireless communication capability enables FL in a wide range of applications such as speech recognition and image classification. Internet of Things (IoT) with billions of devices over the world may provide even larger amount of data while casting further constraints on the computational power and transmission power \cite{mills2019communication}. Therefore, FL has attracted significant interest from both academia and industry, and there is an explosive growth in FL research, such as federated averaging (FedAvg) \cite{mcmahan2017communication}, federated transfer learning (FTL) \cite{yang2019federated}, and FL with differential privacy \cite{wei2020federated}. Among them, model-based FTL stands out as a particularly efficient scheme, because it transfers a well-trained source model into the target task of interest where samples may have different input and output spaces, and thus FTL requires less training samples and shortens the training process \cite{chen2020fedhealth, ju2020federated, yang2020fedsteg}. While the training samples of the target task might be scarce and uploading the parameter updates can be expensive, the open-source big data repositories and deep learning models are becoming prevalent \cite{weiss2016survey}. Through the use of existing datasets or deep learning models based on these datasets that are related to, albeit not exactly the same as, the target input space and output space, transfer learning (TL) is well-known as an attractive solution for various applications, such as text sentiment classification \cite{wang2011heterogeneous, kaya2013transfer, khan2019enhanced}, image classification \cite{duan2012learning, kulis2011you, zhu2011heterogeneous, shaha2018transfer, hussain2018study, han2018new}, human activity classification \cite{harel2010learning, park2016micro, agarwal2021transfer}, software defect classification \cite{ma2012transfer, nam2017heterogeneous}, and multi-language text classification \cite{prettenhofer2010cross, zhou2014heterogeneous, zhou2014hybrid}.

If the clients are mobile devices and the training process of FL is performed over a wireless network such as a wireless local area network (WLAN) or a cellular system, it is highly desired to reduce the amount of uplink payload (i.e., model parameter update), because the available radio spectrum is limited, and the environment is stochastic due to channel fading and interference \cite{chen2021distributed}. Considering the FL applications on IoT devices, the strict budget on computation and uploading power consumption presents additional challenges for complex tasks for which we desire a deeper DNN with massive amount of parameters to train. A large number of aforementioned studies on FL focus on shallow DNNs with three or less layers, while many of the state-of-the-art DNN models have dozens of layers and millions or even billions of trainable parameters to pursue highest accuracy, such as image segmentation \cite{yuan2021segmentation, jain2021semask}, image classification \cite{dai2021coatnet, foret2020sharpness}, object detection \cite{liu2021swin, ghiasi2021simple}, question answering \cite{yamada2020luke}, medical image segmentation \cite{srivastava2021msrf}, and speech recognition \cite{zhang2020pushing}. 

Inspired by the aforementioned challenges on the budget of uplink payload and local computational power, we in this paper propose feature-based federated transfer learning (FbFTL) as a novel approach to perform the model-based transfer FL. Instead of uploading the gradients, we suggest uploading the output and the input of the subset of DNN to be trained, analyze and compare the overall uplink payloads of FL, FTL and FbFTL, and provide the corresponding system design to avoid privacy leakage from the uploaded output. We test this approach by transferring the VGG-16 convolutional neural network (CNN) model \cite{simonyan2014very} trained with ImageNet dataset \cite{deng2009imagenet} to CIFAR-10 dataset \cite{krizhevsky2009learning}, and show that FL, two types of FTL, and FbFTL requires uploading 3216 Tb, 949.5 Tb, 599 Tb, and 6.6 Gb of data, respectively, until performance convergence. Hence, the proposed FbFTL leads to an at least 5 orders of magnitude decrease in upload requirements compared to FL and FTL. We note that, the ITU standard of 5G uplink user experienced data rate is only 50 Mb/s \cite{series2017minimum}, and even the 6G uplink user experienced data rate at Gbit/s level may not sufficiently support such huge uploading requirements of regular FL. Additionally, to our best knowledge, existing works on improving FL efficiency (such as compression, sparsification, quantization, federated distillation, pruning, and over-the-air computation) focus on transmitting gradient updates and have a limited payload reduction of only two orders of magnitude of the original payload on the same CIFAR-10 dataset while also experiencing performance degradation \cite{konevcny2016federatedefficiency, mcmahan2017communication, li2019fedmd, lin2020ensemble, liu2020pruning, aygun2021hierarchical, sun2022time}. Therefore, FbFTL is an economically more viable approach. We also show that FbFTL has significantly lower downlink payload, and transfers the back-propagation computation from the edge devices to the data center, and therefore it is much more friendly in terms of facilitating the training tasks on clients with limited power budget. 

The remainder of the paper is organized as follows. In Section \ref{related}, we introduce the related works on FL and FTL, describe the system design via FedAvg, and demonstrate their uplink payload. In Section \ref{sec:FbFTL}, we propose the FbFTL algorithm, introduce the learning structure and system design, compare its payload with FL and FTL, and analyze random shuffling that preserves the clients' privacy. Subsequently, in the Section \ref{sec:exp}, we evaluate the performance of FbFTL via simulations, and provide comparisons with FL and FTL. Finally, we conclude the paper in Section \ref{sec:con}.

\section{Related Work} \label{related}
In this section, we introduce the related works on FL and FTL, and analyze the least requirements on their successful uplink payload.

\subsection{Federated Learning} \label{subsec:FL}
In this paper, we consider the common FL task for which a PS orchestrates a set $\mathcal{U}$ of $U$ clients to cooperatively train a DNN model. Each client $u$ owns a local dataset $\mathcal{K}_u$ with $K_u$ samples for training, and the $k$th sample $\textbf{s}_{u,k} \in \mathcal{K}_u$ consists of an input vector $\textbf{x}_{u,k} \in \mathbb{R}^{N_0}$ and an output vector $\textbf{y}_{u,k} \in \mathbb{R}^N$. The DNN is a function $f_{\pmb{\theta}}(\textbf{x})$ that maps the input $\textbf{x}$ to an estimated output $\hat{\textbf{y}}$ that aims at estimating $\textbf{y}$ with trainable parameter vector $\pmb{\theta}$. The training process of FL is to update the parameter vector $\pmb{\theta}$ according to the training samples from each client to minimize the expected loss $\mathbb{E} (L(f_{\pmb{\theta}} | \textbf{s}_{test}) )$ on the unseen sample $\textbf{s}_{test}$ which has the same distribution as the training samples. For most DNNs with classification tasks, the output label $\textbf{y}$ is an axis-aligned unit vector with one element equal to 1 indicating the class of this sample, and all others equal to 0. In this case, the loss function is typically the categorical cross-entropy:
\begin{equation} \label{eq:crossentropy}
L(f_{\pmb{\theta}} | \textbf{s}_{u,k}) = - \sum_{n=1}^{N} {\pmb{y}_{u,k}[n] \log{f_{\pmb{\theta}}(\pmb{x}_{u,k})[n]}} .
\end{equation}

In order to minimize the loss and keep the training samples local, the authors in \cite{mcmahan2017communication} proposed an iterative distributed optimization scheme called FedAvg with the following steps:
\begin{enumerate}
    \item The PS initiates trainable parameters $\pmb{\theta}$, and broadcasts the model structure $f$ with non-trainable parameters to each client.
    \item The PS picks a subset of $UC$ clients and broadcasts trainable parameters $\pmb{\theta}$, where $C$ is the fraction of clients selected in this iteration.
    \item Each client uses stochastic gradient descent (SGD) to get the parameter update with respect to each training sample: $\nabla_{\pmb{\theta}}{L(f_{\pmb{\theta}} | \textbf{s}_{u,k})}$.
    \item Each client sends the sum of the updates over all local samples $\textbf{g}_u = \sum_{k=1}^{K_u}{\nabla_{\pmb{\theta}}{L(f_{\pmb{\theta}} | \textbf{s}_{u,k})}}$ and $K_u$ to the PS.
    \item The PS updates the parameter $\pmb{\theta}$ by replacing it with $\pmb{\theta} - \alpha \sum_{u=1}^{U}{ \textbf{g}_u}$ where $\alpha$ is the learning rate that controls the training speed.
    \item Return to step 2) until convergence.
\end{enumerate}

Disregarding the processing failures and transmission failures, we assume that the FL with FedAvg iterates $I^{FL}$ times. Then, all clients upload $\textbf{g}_u$ for $I^{FL} U C$ times during training process. Consider that the DNN consists of $M$ layers with trainable parameters (such as convolutional layer and fully connected layer, i.e., dense layer) and $M'$ layers without trainable parameters (such as pooling layer and residue connection), while the $m$th trainable layer has $T_m$ trainable parameters. Thus, for each trainable parameter, the client uploads one float number of $d$ bits for the corresponding element in the update $\textbf{g}_u$ during each iteration. Therefore, the overall uplink payload to train a DNN via FL with FedAvg is 
\begin{equation} \label{eq:PayloadFL}
P^{FL} = d I^{FL} U C \sum_{m=1}^{M}{T_m} \  \text{bits},
\end{equation}
and we will set this as a benchmark to compare with three other training methods in the remainder of this paper.

\subsection{Federated Transfer Learning} \label{subsec:FTL}
TL has been shown to be a successful learning technique that leverages the knowledge from a different domain to improve the performance in the target domain, and FTL extends the traditional TL to the privacy-preserving distributed machine learning paradigm to mitigate the challenges arising from the scarcity of spectrum and training data in wireless FL. In this paper, we consider FTL to be performed in the same fashion as in the previous subsection, where one PS orchestrates $U$ clients. Based on the difference between the domains, FTL can be divided into three different categories: instance-based FTL, feature-based FTL, and model-based FTL \cite{yang2019federated, 5640675}. The former two types of FTL assume similarity in the distribution of the input and output, and hence we in this paper focus on the model-based FTL which only assumes similarity in the functionality to extract a high-dimensional description from the input data. 

\begin{multicols}{2}
\end{multicols}
\begin{figure*}[h]
  \centering
  \includegraphics[width=0.8\textwidth]{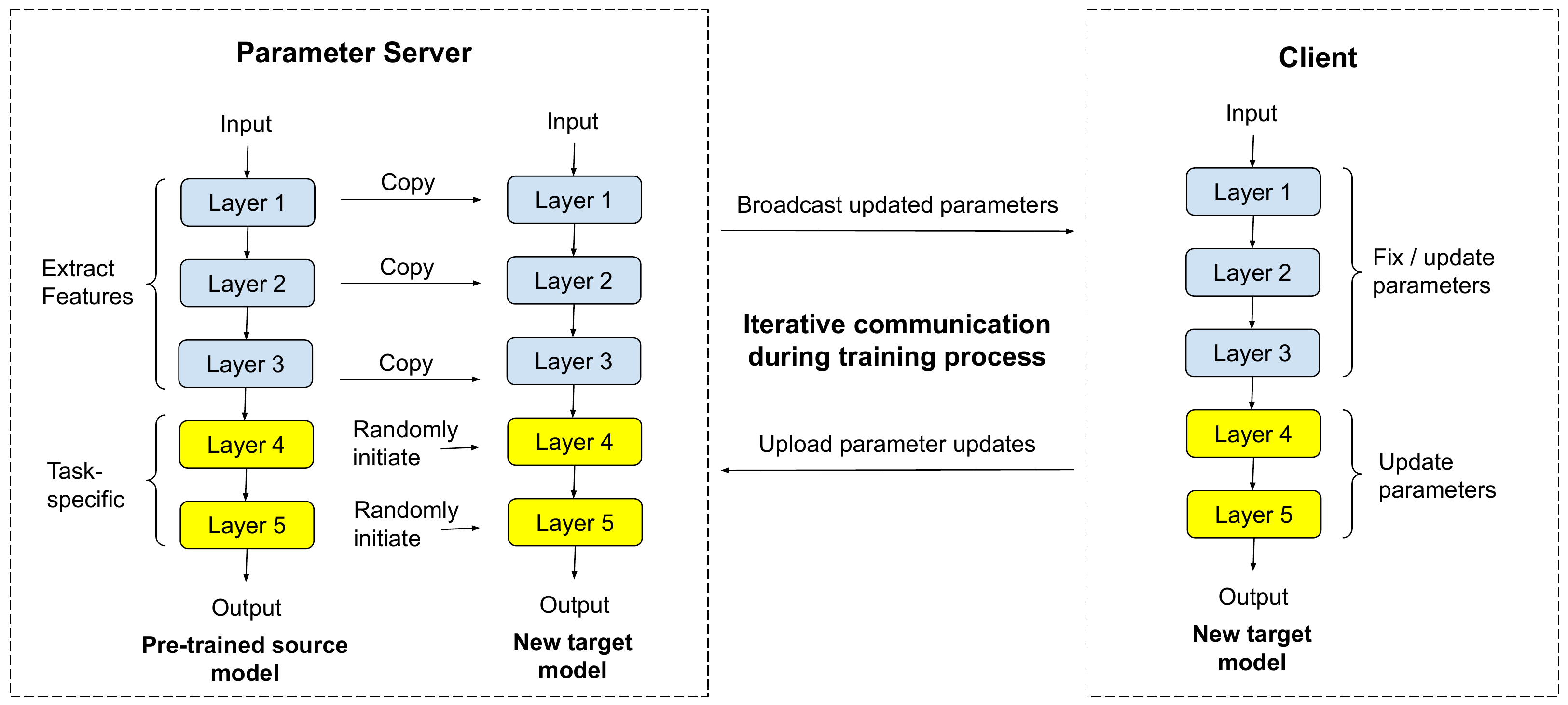}
  \caption{Diagram of the iterative training process of model-based federated transfer learning.}
  \label{fig:FTLdiagram}
\end{figure*}
\begin{multicols}{2}
\end{multicols}
As shown in Fig. \ref{fig:FTLdiagram}, we consider FTL on a pre-trained DNN model based on a different open-source dataset with similar but not the same task as the source model to reduce the number of iterations during training. Many of the DNNs can be divided into two parts, the first part extracts the high-dimensional features from the sample input data with general information, and the second part performs operations for a specific task such as classification and is less possible to transfer into a new task. To transfer the knowledge of the source model into a new task before the training starts, the feature extraction part of the source model is directly copied into the new target model, while the task-specific part is randomly initiated. One way to perform FTL is to retrain all parameters with the new data. However, considering the number of parameters to train, one of the most common ways to perform FTL with the aim to minimize the training time and data requirement is to fix the feature extraction part and simply update the task-specific part. Specifically, we select one fully connected layer $m_c$ (which is Layer 4 in Fig. \ref{fig:FTLdiagram}), close to the output without paralleling path (such as residue connection), and randomly initiate all trainable parameters of layers $m_c, m_c+1, \ldots, M$ (which are Layer 4 and Layer 5 in Fig. \ref{fig:FTLdiagram}). During the distributed training process, all copied parameters are fixed, and we only update the parameters of layers $m_c, \ldots, M$.

Similar to the previous sub-section on FL, we analyze the least requirements on successful uplink payload for FTL until convergence. For FTL that updates all parameters, assuming that FTL with FedAvg is iterated $I^{FTL}_{f}$ times, the overall uplink payload required for training is 
\begin{equation} \label{eq:PayloadFTLfull}
P^{FTL}_{f} = d I^{FTL}_{f} U C \sum_{m=1}^{M}{T_m} \  \text{bits}.
\end{equation}
On the other hand, assuming FTL with FedAvg that only updates the task-specific part with $I^{FTL}_{c}$ iterations and assuming that each sum update $\textbf{g}_u$ consists of $\sum_{m=m_c}^{M}{T_m}$ trainable parameters, the overall uplink payload for training is 
\begin{equation} \label{eq:PayloadFTLclassify}
P^{FTL}_{c} = d I^{FTL}_{c} U C \sum_{m=m_c}^{M}{T_m} \  \text{bits}.
\end{equation}
Typically, training from scratch requires more training samples, and thus we have $I^{FTL}_{c}UC \approx I^{FTL}_{f}UC < I^{FL}UC$. For most of the models, the majority of the DNN structure is dedicated to the feature extraction, and obviously $\sum_{m=m_c}^{M}{T_m} < \sum_{m=1}^{M}{T_m}$. Therefore, 
\begin{equation} \label{eq:compareFLFTL}
P^{FTL}_{c} < P^{FTL}_{f} < P^{FL}.
\end{equation}

\section{Feature-based Federated Transfer Learning}\label{sec:FbFTL}
In this section, we propose the FbFTL framework that requires significantly less uplink payload than FL and FTL, and this is achieved by uploading the extracted features and outputs instead of the parameter updates. To prevent privacy leakage from outputs, we also propose an uploading design to conceal the dependency between the client address and the output data.

\subsection{Learning Structure} \label{subsec:FbFTLLearn}
\begin{multicols}{2}
\end{multicols}
\begin{figure*}[h]
  \centering
  \includegraphics[width=0.8\textwidth]{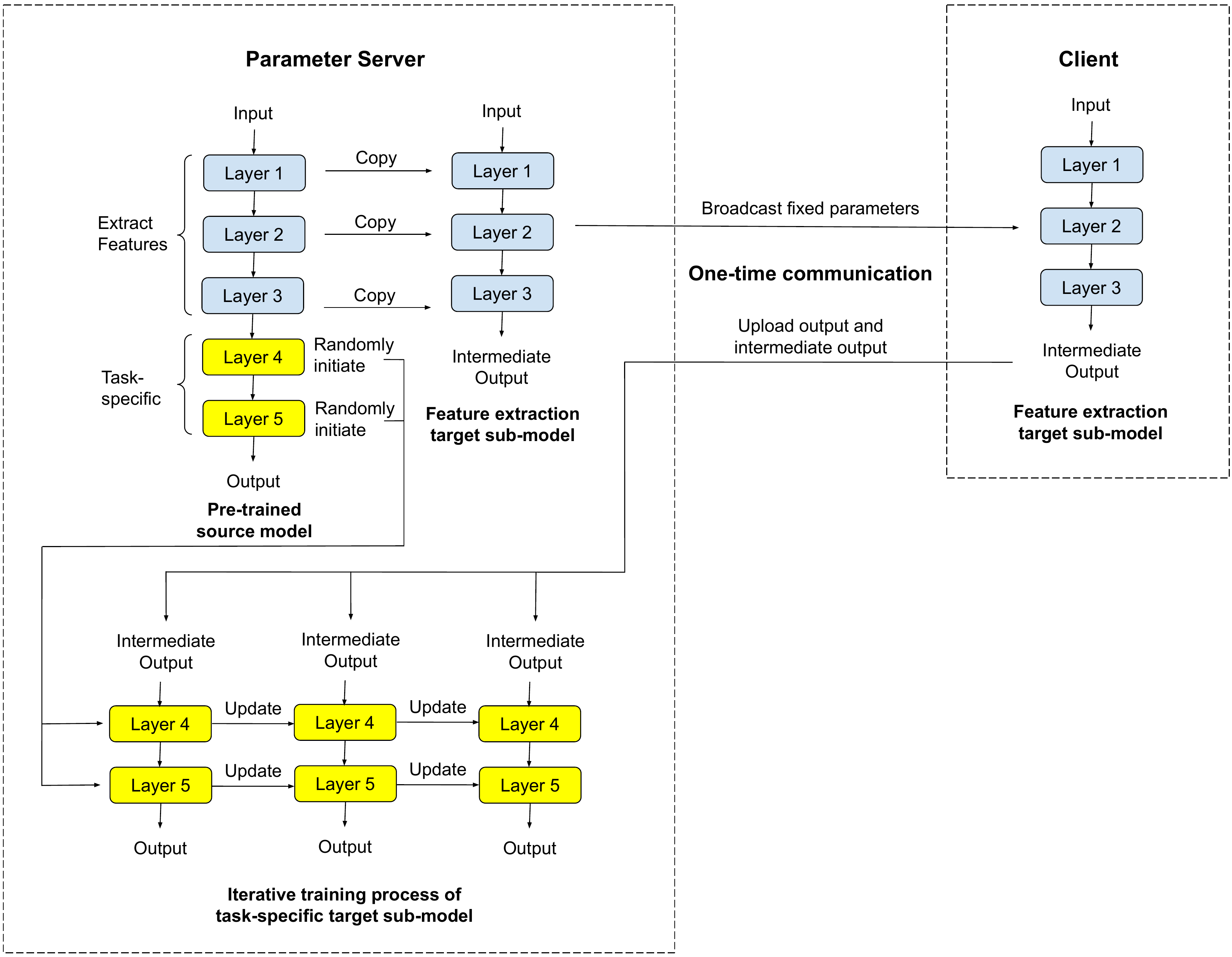}
  \caption{Diagram of the training process of feature-based federated transfer learning, which reduces to one-time communication of the output and the intermediate output (i.e., extracted features).}
  \label{fig:FbFTLdiagram}
\end{figure*}
\begin{multicols}{2}
\end{multicols}
As shown in Fig. \ref{fig:FbFTLdiagram}, in FbFTL we consider the model-based TL on a source DNN model $f'$ which is pre-trained on a different task. We pick one fully connected layer $m_c$ without paralleling path as the cut layer, and divide the new target model $f \leftarrow f'$ into two parts. Those layers before layer $m_c$ (which is Layer 4 in Fig. \ref{fig:FbFTLdiagram}) are regarded as the feature extraction sub-model $f^1_{\pmb{\theta}^1}$, and the parameters $\pmb{\theta}^1$ for the new target model $f$ are copied from the pre-trained source model $f'$ and fixed without any further update. The other layers are regarded as the task-specific sub-model $f^2_{\pmb{\theta}^2}$, and all trainable parameters $\pmb{\theta}^2$ in the new target model $f$ are randomly initiated for training with the $u$th client's $k$th training sample $\textbf{s}_{u,k} = \{ \textbf{x}_{u,k}, \textbf{y}_{u,k}\}$ for all $u$ and $k$. 

The forward pass of the full model $f_{\pmb{\theta}}(\textbf{x}_{u,k}) = f^2_{\pmb{\theta}^2}(f^1_{\pmb{\theta}^1}(\textbf{x}_{u,k}))$ maps the input $\textbf{x}_{u,k}$ to the estimated output $\hat{\textbf{y}}_{u,k}$, and we note that the output of the feature extraction sub-model $\textbf{z}_{u,k} = f^1_{\pmb{\theta}^1}(\textbf{x}_{u,k})$ is also the input of the task-specific sub-model $f^2_{\pmb{\theta}^2}$. Since we only desire to train the task-specific sub-model, there may only be forward path from input $\textbf{x}_{u,k}$ that generates features $\textbf{z}_{u,k}$ at each client and sends to the PS, and the gradient back-propagation is iteratively performed at the PS without any feedback to each client in FbFTL. In contrast, the parameter update of FL and FTL based on the gradients highly depends on the parameters in the current training iteration, and the same data sample may generate different parameter updates in different iterations during the training process. Therefore in FL and FTL, we either require many more training samples, or need to perform uploading updates multiple times for the same sample. However in FbFTL, each client only needs to upload the intermediate output $\textbf{z}_{u,k}$ and output $\textbf{y}_{u,k}$ once, instead of iteratively uploading the gradients, and the PS may regard these as the input and the output of the training sample for the task-specific sub-model $f^2_{\pmb{\theta}^2}$. Such pairs of samples are not correlated with the model parameters $\pmb{\theta}^2$, and therefore they can be used at different training iterations without downloading or uploading anything again. We provide the steps of the FbFTL algorithm in Algorithm \ref{alg:FbFTL} below.

\begin{algorithm}
\caption{FbFTL}
	\label{alg:FbFTL}
	\begin{algorithmic}
	    \State{PS copies fixed sub-model $f^1_{\pmb{\theta}^1}$ from pre-trained source model and randomly initiates trainable sub-model $f^2_{\pmb{\theta}^2}$}.
	    \State{PS broadcasts $f^1_{\pmb{\theta}^1}$ to each client $u \in \mathcal{U}$}
	    \State{\textbf{Clients execute:}}
	    \For{client $u \in \mathcal{U}$}
	        \For{sample $\textbf{s}_{u,k} \in \mathcal{K}_u$}
	            \State {Upload $\textbf{z}_{u,k} = f^1_{\pmb{\theta}^1}(\textbf{x}_{u,k})$ and $\textbf{y}_{u,k}$ to PS}
	        \EndFor
	    \EndFor
	    \State{\textbf{PS executes:}}
	    \While{$not\ converge$}{
	        \For{mini-batch $b$ of pairs $\{u, k\} \in \mathcal{U} \times \mathcal{K}_u$}
	            \State{$\pmb{\theta}^2 \leftarrow \pmb{\theta}^2 - \alpha \sum_{\{u, k\} \in b}{\nabla_{\pmb{\theta}^2}{L(f_{\pmb{\theta}^2} | \textbf{z}_{u,k}, \textbf{y}_{u,k})}}$}
	        \EndFor
	    \EndWhile}
	\end{algorithmic}
\end{algorithm}

We note that by uploading the training samples of task-specific sub-model instead of the direct gradient updates, FbFTL also has additional critical benefits in improving the training performance in practice, such as hyper-parameter fine-tuning, dataset balancing, and enabling flexible training batch size selection, as described in detail next.

Specifically, the most important benefit is that FbFTL enables iterative fine-tuning of the SGD optimizer hyper-parameters such as the learning rate. To obtain the optimized DNN, one needs to find the optimized hyper-parameters that provide the best convergence performance. To train a DNN via FbFTL, one can train the model from scratch for several times to determine the best hyper-parameters without additional communication with the clients. However for FL and FTL, the entire online training process has to be run several times, costing much more than the ideal uplink payload $P$. 

Another benefit of FbFTL is the dataset balancing. If the overall dataset is imbalanced and hence the samples with certain types of output appears much more frequently than those of other outputs, it is hard to distinguish for FL and FTL via gradient updates, and such imbalanced data distribution could significantly
degrade FL performance \cite{zhao2018federated, park2019wireless}. However, FbFTL with direct output information enables techniques, such as re-sampling specific classes or merging near-identical classes, to improve dataset imbalance.

One more benefit of FbFTL is to lift the constraint of $UC$ and $K_u$ on the training batch size. To avoid over-fitting to the training set of data, one needs to validate the performance on a separate validation set of data to identify the optimal number of training iterations to stop training and conclude the final model. It is straightforward for FbFTL to divide the obtained dataset into training and validation subsets, but it calls for extra effort for the communication system to meticulously operate the training process and validation process with the desired order and number of samples. Additionally for the training process, due to the broadcast nature of the downlink in wireless FL and FTL, all selected clients in the same communication iteration receives the same parameters, so each SGD mini-batch has a larger size than $\sum_u^{UC} K_u$ and is on the order of given clients' samples. However, FbFTL may choose any rational size of SGD mini-batch without the constraints of the communication system, and can reshuffle the data in each training iteration.

\subsection{Payload Analysis} \label{subsec:FbFTLpayload}
Except the one-time communication benefit of FbFTL, we note that the amount of data in the single upload batch in FbFTL can be much less than that of each sample in the upload batch of FL and FTL. For each fully connected layer $m$ with bias in the model, we denote the number of input nodes as $N^{-}_{m}$ and the number of output nodes as $N^{+}_{m}$. Then, the number of trainable parameters in this layer is $T_m = N^{+}_{m} (N^{-}_{m} + 1)$. Compared to the dimension of $N^{-}_{m_c}$, the amount of information for $\textbf{y}_{u,k}$ which is typically a single integer is negligible. Note that FbFTL requires calculating gradients at the PS, and hence FedAvg is not applicable and each sample has to be uploaded separately. Therefore for FbFTL, the uplink payload for each sample is $d N^{-}_{m_c}$, and the overall successful uplink payload required for training is 
\begin{equation} \label{eq:PayloadFbFTL}
P^{FbFTL} = d \sum_{u = 1}^{U}{K_u} N^{-}_{m_c} \  \text{bits}.
\end{equation}
Compared with the uplink payload of FTL with FedAvg that only updates the task-specific sub-model in (\ref{eq:PayloadFTLclassify}), the ratio of each upload for single sample between FTL and FbFTL is 
\begin{multline} \label{eq:FTL/FbFTL_single}
\begin{aligned} 
\frac{d \sum_{m=m_c}^{M}{T_m}}{d N^{-}_{m_c}} & = \frac{T_{m_c} + \sum_{m=m_c+1}^{M}{T_m}}{N^{-}_{m_c}} \\
& = \frac{N^{+}_{m_c} (N^{-}_{m_c} + 1) + \sum_{m=m_c+1}^{M}{T_m}}{N^{-}_{m_c}} \\
& > N^{+}_{m_c} .
\end{aligned}
\end{multline}
Therefore, 
\begin{multline} \label{eq:FTL/FbFTL_all}
\begin{aligned} 
\frac{P^{FTL}_{c}}{P^{FbFTL}} & = \frac{d I^{FTL}_{c} U C \sum_{m=m_c}^{M}{T_m}}{d \sum_{u=1}^{U}{K_u} N^{-}_{m_c}} \\
& > \frac{I^{FTL}_{c} U C}{\sum_{u=1}^{U}{K_u}} N^{+}_{m_c}. \\
\end{aligned}
\end{multline}
The number of extracted features for many state-of-the-art models is larger than $10^3$, and TL usually requires a relatively deeper task-specific sub-model, and therefore $N^{+}_{m_c}$ can be large. For the cross-device federated learning, the clients do not obtain huge local datasets, $\sum_{u=1}^{U}{K_u} < I^{FTL}_{c} U C$. Therefore, we have $P^{FbFTL} \ll P^{FTL}_{c}$. We note that FbFTL and FTL updating task-specific sub-model have the same performance at every iteration because the only difference between the two methods is the communication system structure but not the numerical process to generate the gradient updates. Combining this characterization with the conclusion in (\ref{eq:compareFLFTL}), we have 
\begin{equation} \label{eq:compareFLFTLFbFTL}
P^{FbFTL} \ll P^{FTL}_{c} < P^{FTL}_{f} < P^{FL} ,
\end{equation}
and therefore we expect extremely less uplink payload in FbFTL compared to FTL and FL.

We also note that FbFTL has the least downlink broadcast payload and the least local computation compared to FL and FTL. On the one hand, for each method of FL, FTL updating all parameters, FTL updating task-specific sub-model, and FbFTL, the overall downlink broadcast payloads, respectively, are
\begin{equation} \label{eq:broadcastFL}
D^{FL} = d I^{FL} \sum_{m=1}^{M}{T_m} \  \text{bits},
\end{equation}
\begin{equation} \label{eq:broadcastFTLf}
D^{FTL}_{f} = d I^{FTL}_{f} \sum_{m=1}^{M}{T_m} \  \text{bits},
\end{equation}
\begin{equation} \label{eq:broadcastFTLc}
D^{FTL}_{c} = d I^{FTL}_{c} \sum_{m=1}^{M}{T_m} \  \text{bits},
\end{equation}
\begin{equation} \label{eq:broadcastFbFTL}
D^{FbFTL} = d \sum_{m=1}^{m_c-1}{T_m} \  \text{bits}.
\end{equation}
On the other hand, for FL and FTL, each client must complete one full forward pass and one backward pass of the parameters to be trained for each sample at each iteration. However for FbFTL, each client only needs to complete the forward pass of feature extraction sub-model once for each sample, and all other computations are transferred to the PS. We note that at the PS, stable electricity supply and advanced processors with much higher efficiency such as graphics processing units (GPUs) can be deployed to facilitate these computations. And, this transfer of computations to the PS makes FbFTL especially advantageous when users and devices have limitations on their computational capabilities and power (e.g., as in IoT devices).

\subsection{Privacy Analysis} \label{subsec:FbFTLprivacy}
Finally, we analyze how random shuffling preserves the clients' privacy while uploading the intermediate output $\textbf{z}$ and the overall output $\textbf{y}$. The intermediate output $\textbf{z}$ is also referred to as the smashed data in split learning \cite{gupta2018distributed, vepakomma2018split}, and cannot be transformed back to the input $\textbf{x}$ due to the non-linearity of the activation functions in each layer and will protect clients' privacy as the gradient updates do in FL and FTL. However, the output $\textbf{y}$ potentially conveys the clients' private information to a minor extent. Due to this, we propose a specific system design to conceal the relationship between the client's address and uploaded content, and thus mix the output information from all clients to eliminate the privacy leakage.

To validate this approach, we analyze the private information leakage of a specific batch of all samples from one client to a potential adversary without client address to distinguish the correspondence. According to \cite{longpr2017entropy}, the privacy loss of a query is the difference between the amount of privacy in the original setting and that after query, and in our case, the amount of privacy can be described by the uncertainty from the adversary's perspective. Subsequently, we in this section conduct the analysis on how much the random shuffling of batches from a large number of clients reduces the amount of privacy loss.

\begin{figure}
	\centering
	\includegraphics[width=0.95\linewidth]{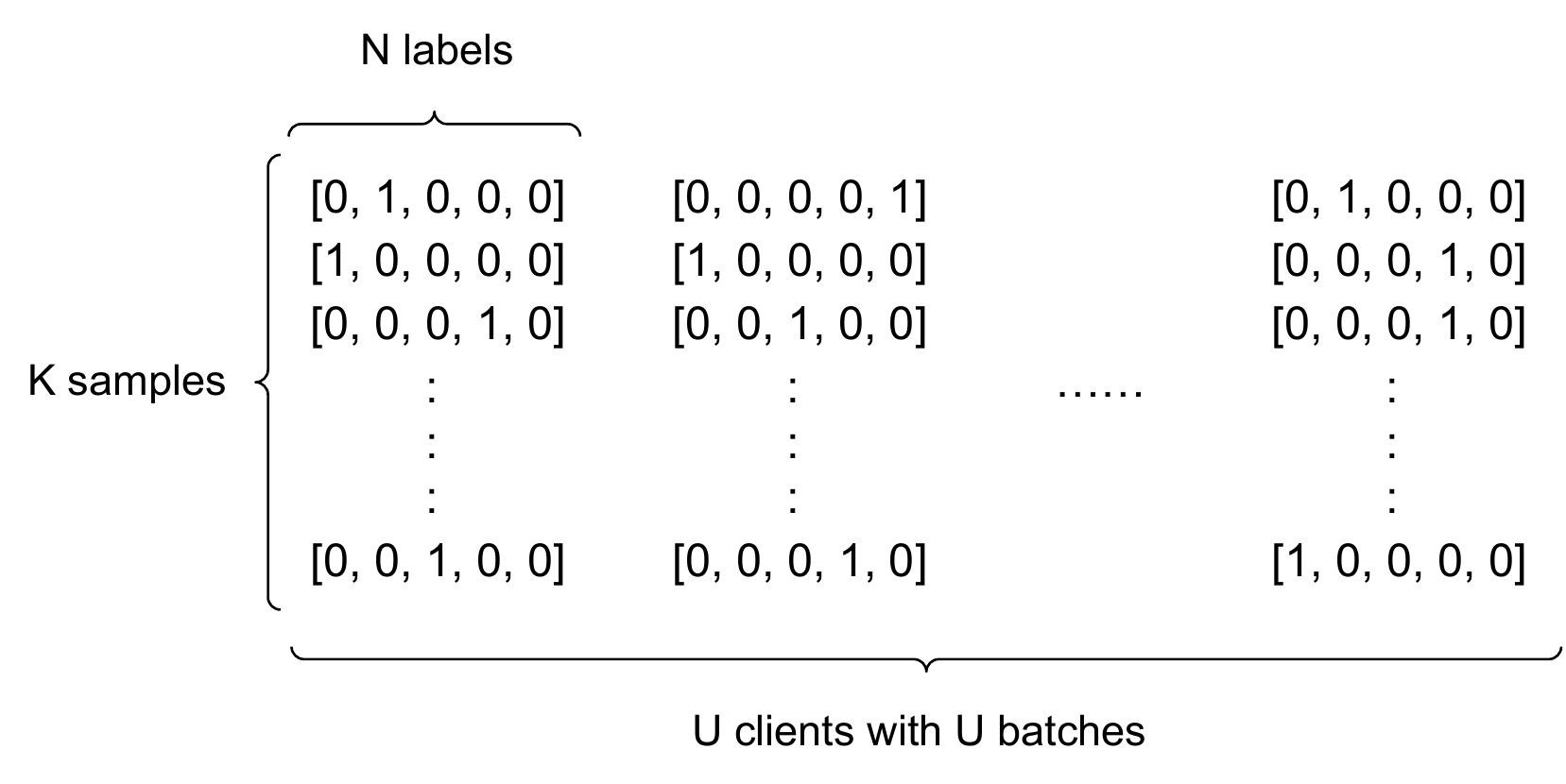}
	\caption{Diagram of every sample label $\textbf{y}_{u,k} \in [0,1]^N$ in batches. }
	\label{fig:batch}
\end{figure}

As shown in Fig. \ref{fig:batch}, we consider a common learning task setting where the output $\textbf{y} \in [0,1]^N$ is an axis-aligned unit vector with one element having a value of 1 indicating the label (manual choice for the sample by the user who owns the client), and all others equal to 0. We assume that each sample $\{\textbf{x}_{u,k}, \textbf{z}_{u,k}, \textbf{y}_{u,k}\}$ is independent and identically distributed (i.i.d.). Each client $u \in \mathcal{U}$ transmits one batch with $K_u = K$ samples, the same number of samples including $\textbf{y}_{u,k}$. Therefore for this client, there are $N^{K}$ different possible ordered batches in total. 

Assuming the order of choices in the batch does not hold information, we use $\mathcal{B}$ to denote the set of possible batches without order from client $u$, and the privacy information of the real batch $b$ from client $u$ is the sum vector $\sum^{K}_{k=1} \textbf{y}_{u,k} = [n^y_{1,b}, n^y_{2,b}, \ldots, n^y_{N,b}]$, and $\sum^{N}_{a=1} n^y_{a,b} = K$. Subsequently, the number of possible batches without order is the combination with replacement of $N$ items taken $K$ times, and is given by 
\begin{equation} \label{eq:size_B}
|\mathcal{B}| = \binom{N + K - 1}{K} .
\end{equation}
We denote the index of client $u$'s batch as a random variable $B$, and the index of the uploaded batch as $B = b \in \mathcal{B}$. 
We denote the type of the label $\textbf{y}_{u,k}$'s distribution as $Y$. Typically to achieve better performance, in machine learning applications we desire a uniform distribution over different labels, and we denote it as $Y=y_0=[\frac{1}{N}, \frac{1}{N}, \ldots, \frac{1}{N}]$. 

To evaluate the adversary's knowledge on one given batch $B=b$ from the given sample distribution $Y=y$, we consider the optimally presumed distribution $\textbf{P}^a_{B}$ of the batch $B$ from the adversary's perspective (instead of the averaged probability of random experiments) to generate the adversary's uncertainty $H^a(B)$ of the given batch (instead of the entropy of random batches). 
A strong adversary with the ability to steal and decode the uploaded data must understand the structure of the DNN and is able to decode the labels of the $K$ samples in a batch, and hence the adversary is likely to have the prior knowledge of the general distribution of batches from a large number of clients. 

In the case of $Y=y_0$ with samples uniformly distributed, the distribution of ordered batches from the adversary's perspective is also uniform with probability $1/N^{K}$. Each batch index $b \in \mathcal{B}$ without order corresponds to $K! / {\prod^N_{a=1} n^y_{a,b}!}$ different batch indices with order, so the adversary's presumed probability distribution of batches without order is $\textbf{P}^a_{B|y_0} = [p^a_{0,1}, p^a_{0,2}, \ldots, p^a_{0,|\mathcal{B}|}]$, where
\begin{equation} \label{eq:dist_without_order}
p^a_{0,b} = \frac{K!}{N^{K} \prod^N_{a=1} n^y_{a,b}!} . 
\end{equation}
Therefore from the adversary's perspective, the uncertainty of the batch from client $u$ with known format and size but without prior knowledge about the content is the entropy
\begin{equation} \label{eq:entropy_u}
H^a(B\ |\ Y=y_0) = - \sum^{|\mathcal{B}|}_{b'=1} p^a_{0,b'} \log_2{p^a_{0,b'}}  .
\end{equation}

In the case where the true distribution is $Y=y$, we denote the adversary's presumed distribution of the batch from client $u$ as  
\begin{equation} \label{eq:dist_BgivenY}
\textbf{P}^a_{B|y} \triangleq \textbf{P}^a\{ B\ |\ Y=y\} = [p^a_{y,1}, p^a_{y,2}, \ldots, p^a_{y,|\mathcal{B}|}] ,
\end{equation}
and the corresponding uncertainty at the adversary in the original setting before query is
\begin{equation} \label{eq:entropy_ugivenY_extreme}
H^a(B\ |\ Y=y) = - \sum^{|\mathcal{B}|}_{b'=1} p^a_{y,b'} \log_2{p^a_{y,b'}}   .
\end{equation}

In the worst case for FbFTL, the strongest query is the process that the adversary acquires all batches without clients' addresses. Since the batches are randomly shuffled without the knowledge of clients' addresses, which batch out of $U$ batches corresponds to which client has a uniform distribution from the adversary's perspective. We use $S$ to denote the number of shuffled batches with different types $b$, and $s(b)$ is the count for each type. Therefore, the adversary's presumed probability distribution of $B$ given this information $S=s$ but without prior knowledge on $Y$ is equal to the frequency of each batch type in $\mathcal{B}$, which is
\begin{equation} \label{eq:dist_BgivenS}
\textbf{P}^a_{B|s} \triangleq \textbf{P}^a\{B\ |\ S=s\} = \bigg[ \frac{s(1)}{U}, \frac{s(2)}{U}, \ldots, \frac{s(|\mathcal{B}|)}{U}\bigg] .
\end{equation}
Therefore, the adversary's uncertainty given the shuffled data $S=s$ is
\begin{equation} \label{eq:entropy_shuffled}
H^a(B\ |\ S=s) = - \sum^{|\mathcal{B}|}_{b'=1} \frac{s(b')}{U} \log_2{\frac{s(b')}{U}} .
\end{equation}

Subsequently in the case where the adversary has prior knowledge $Y=y$ and acquired $S=s$ via the query, we denote the adversary's presumed distribution as $\textbf{P}^a(B\ |\ Y=y, S=s)$, and its uncertainty on client $u$'s batch $b$ as $H^a(B\ |\ Y=y, S=s)$. Then, we have
\begin{multline} \label{eq:ys_equal_s}
\begin{aligned} 
& \textbf{P}^a(B=b|Y=y,S=s) \\
= & \frac{\textbf{P}^a(S=s|B=b,Y=y) \textbf{P}^a(B=b|Y=y)}{\textbf{P}^a(S=s|Y=y)} \\
= & {\binom{U-1}{s(b)-1}} / {\binom{U}{s(b)}} \\
= & \frac{s(b)}{U} \\
= & \textbf{P}^a\{B=b|S=s\} .\\
\end{aligned}
\end{multline}
This result indicates that the adversary's presumed distribution depends only on the frequency of each batch type.

Therefore, the privacy leakage when the samples are shuffled is
\begin{multline} \label{eq:leakage_shuffled}
\begin{aligned} 
& H^a(B\ |\ Y=y) - H^a(B\ |\ Y=y, S=s) \\
= & H^a(B\ |\ Y=y) - H^a(B\ |\ S=s)  . \\
\end{aligned}
\end{multline}
Since the number of shuffled batches $s$ of certain type is the count over $U$ sampled batches from the true distribution $y$, when the number of clients $U$ is small, it is likely that $\textbf{P}^a_{B|y} \neq \textbf{P}^a_{B|s}$ and the adversary obtains additional information from getting the shuffled data $S=s$. However when $U$ is large, $\lim\limits_{U\to\infty} \textbf{P}^a_{B|s} = \textbf{P}^a_{B|y}$ by the law of large numbers, and the privacy leakage converges to 0. Typically, federated learning has a large number of clients (i.e., $U$ is typically large) while $N$ and $K$ are limited, and the privacy leakage with shuffled output is negligible. 

The superiority of shuffling is also proved in \cite{erlingsson2019amplification} in terms of differential privacy, where the random shuffled data guarantee a factor $\sqrt{U}$ of privacy cost reduction in the worst case. 

In order to segregate the client's Internet Protocol (IP) address from the uploaded content including the output $\textbf{y}$, we consider the setting in which FbFTL is performed by a commercial corporation via the cellular network and training data from clients is acquired via multiple base stations. Typically, the uploaded content is encrypted and passed from the base station to the core network and then issued to the commercial corporation. The commercial corporation may formulate its own encryption scheme that only shares with the clients in addition to the encryption of the communication operator, so that it is impossible for the communication operator with the IP addresses to extract the contents. When the communication operator passes the contents to the commercial corporation, it should randomly reorder each content from different clients and delete the corresponding IP addresses, so that it is impossible for the commercial corporation to obtain the IP addresses while it has access to the contents. Therefore, no one other than the client may obtain both its IP address and its content, and we successfully prevent the PS from locating each client.

\section{Experimental Results}\label{sec:exp}

In this section, we apply FL, FTL and FbFTL on VGG-16 CNN model \cite{simonyan2014very} and transfer the knowledge learned from ImageNet dataset \cite{deng2009imagenet} to CIFAR-10 dataset \cite{krizhevsky2009learning}. 

\begin{figure}
	\centering
	\includegraphics[width=1\linewidth]{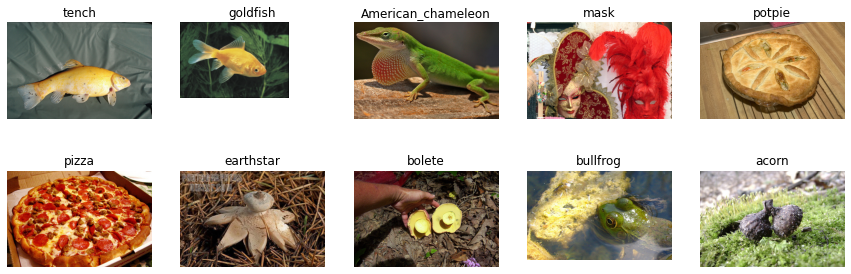}
	\caption{ImageNet samples with labels. }
	\label{fig:imagenet}
\end{figure}

ImageNet is a large virtual database with more than 14 million images with hand-annotated labels (classification types or desired outputs for training), and the pre-trained source model that we use for TL is developed on the ImageNet large scale visual recognition challenge 2012 (ILSVRC2012, \cite{ILSVRC15}) with images from 1000 categories. As an example, we show 10 samples with their labels in Fig. \ref{fig:imagenet}.

\begin{figure}
	\centering
	\includegraphics[width=0.8\linewidth]{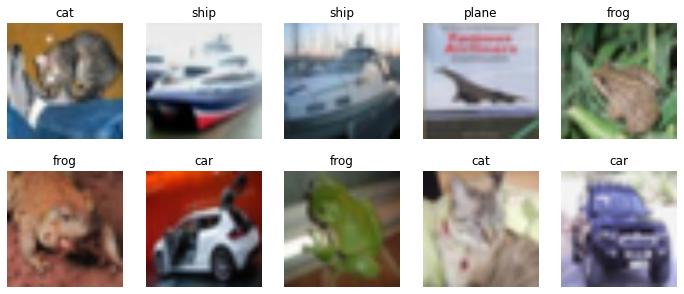}
	\caption{CIFAR-10 samples with labels. }
	\label{fig:cifar10}
\end{figure}

CIFAR-10 is another database of 60000 images with $N = 10$ different labels, and we use 50000 images for training and 10000 images for testing. Fig. \ref{fig:cifar10} shows the first 10 samples with their labels, and we note that these samples have simple labels and naturally seem blurred compared to those in ImageNet because images in CIFAR-10 have a lower dimension. 

\begin{figure}
	\centering
	\includegraphics[width=0.6\linewidth]{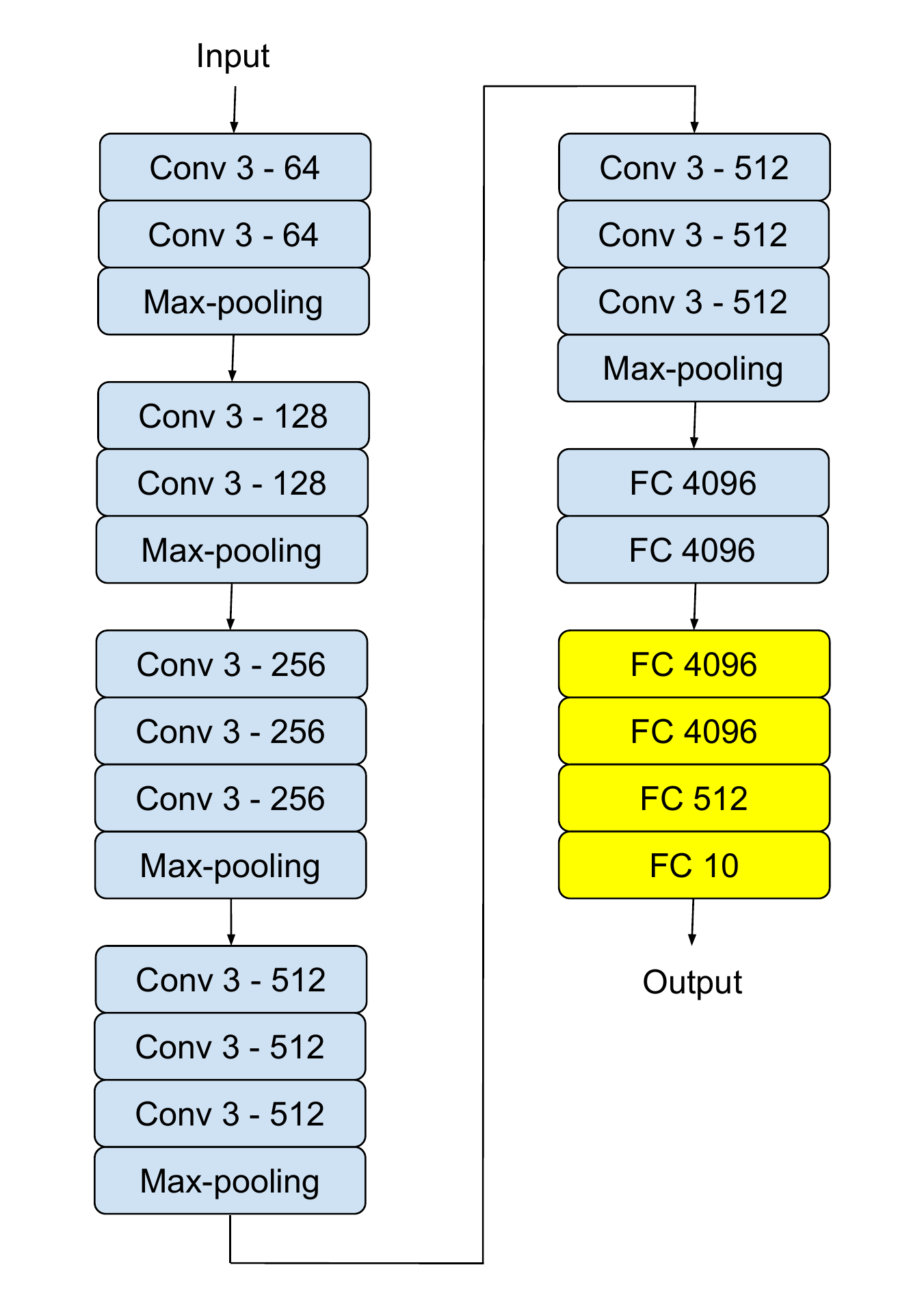}
	\caption{Diagram of the VGG-16 model for training on CIFAR-10 dataset. The convolutional layers are denoted as "Conv $\{$receptive field size$\}$ - $\{$number of output channels$\}$", and the fully connected layers are denoted as "FC $\{$output size$\}$". For brevity, the ReLU activation function and dropout are not shown. }
	\label{fig:VGG16}
\end{figure}

In Fig. \ref{fig:VGG16}, we show the structure of VGG-16 that we use for training on CIFAR-10. For TL, the first half of the layers marked blue are considered as the feature extraction sub-model $f^1_{\pmb{\theta}^1}$ and are copied from that trained on ImageNet, and the latter part marked yellow is considered as the task-specific sub-model $f^2_{\pmb{\theta}^2}$ and is randomly initiated. For FbFTL, the dimension of the intermediate output is $N^{-}_{m_c} = 4096$.

To train the model on CIFAR-10, we use Nvidia GeForce GPU with CUDA to simulate the algorithms with PyTorch \cite{pytorch}. We assume that there are $U=6250$ clients in total, each iteration takes a fraction $C=1.28 \times 10^{-3}$ of all clients, and each batch contains $K_u=8$ samples. In the two most commonly used deep learning tools TensorFlow (including Keras) \cite{tensorflow2015-whitepaper} and PyTorch, the default data type of each number has 32 bits and therefore $d=32$ bits. The learning rate is $10^{-2}$, the momentum of the optimizer is $0.9$, and the L2 penalty is $5 \times 10^{-4}$.

In Table \ref{tab:exp}, we compare the performances of FL, FTL updating full model, FTL updating task-specific sub-model, and FbFTL. As we have analyzed, the first three algorithms require $IUC$ successfully uploaded batches, while FbFTL only requires $\sum_{u=1}^{U}{K_u}$ batches. Also, the first three algorithms require uploading $\sum_{m=1}^{M}{T_m}$, $\sum_{m=1}^{M}{T_m}$, and $\sum_{m=m_c}^{M}{T_m}$ parameters, respectively, in each batch, while FbFTL only requires uploading $N^{-}_{m_c}$ in each batch. In the third row of the table, we note that FbFTL has a reduction of four orders of magnitude (i.e., reduction by $10^{-4}$) in uplink payload per batch compared to those of the other algorithms for each client, and in the fourth row, it achieves five orders of magnitude reduction (i.e., reduction by $10^{-5}$) in total uplink payload compared to those of the other algorithms during training. We note that the ITU standard of 5G uplink user experienced data rate is only 50 Mb/s\cite{series2017minimum}, and FbFTL is obviously the most economically prominent approach. There is also a huge reduction on the overall downlink payload by applying FbFTL. At the same time, we also note that FTL updating task-specific sub-model and FbFTL only update a small portion of parameters and have slightly lower performance in terms of test accuracy, which is the trade-off that arises with the payload reduction. If we are to train a larger model on a more complex task or the size of training dataset is more restricted, the difference between the payload can be even larger. With the same task but 10\% of the training samples, FbFTL achieves over 80\% accuracy with less than 1 Gb total uplink payload.

\begin{table}
	\fontsize{7}{8.5}
	\selectfont
	\begin{center}
		\caption{Comparison between FL, FTL updating full model, FTL updating task-specific sub-model and FbFTL}
		\label{tab:exp}
		\begin{tabular}{ | l || l | l | l | l | }
			\hline
			 & FL & FTL$_f$ & FTL$_c$ & FbFTL \\ \hline\hline
			upload batches & 656250 & 193750 & 525000 & 50000  \\ \hline
			upload parameters per batch & 153144650 & 153144650 & 35665418 & 4096  \\ \hline
			uplink payload per batch & \textbf{4.9 Gb} & \textbf{4.9 Gb} & \textbf{1.1 Gb} & \textbf{131 Kb}  \\ \hline
			total uplink payload $P$ & \textbf{3216 Tb} & \textbf{949 Tb} & \textbf{599 Tb} & \textbf{6.6 Gb}  \\ \hline
			total downlink payload $D$ & 402 Tb & 253 Tb & 322 Tb & 3.8 Gb  \\ \hline
			test accuracy & 89.1\% & 91.68\% & 85.59\% & 85.59\%  \\ \hline
		\end{tabular}
	\end{center}
\end{table}

\begin{figure}
	\centering
	\includegraphics[width=0.8\linewidth]{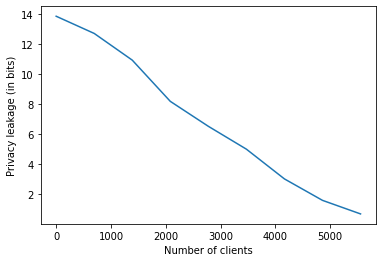}
	\caption{Privacy leakage of each batch for FbFTL in bits against the number of clients. }
	\label{fig:leakage}
\end{figure}

In Fig. \ref{fig:leakage}, we show the amount of privacy leakage of a certain batch described in (\ref{eq:leakage_shuffled}) in the settings with different number of clients $U$. The curve starts with only one client, and in that case all the information (13.86 bits) is leaked to the adversary. As the number of participating clients increases, the privacy leakage drops fast in our considered experiment setting (with 6250 clients). In comparison, the dimension of the input information for each batch is $8 \times 32^2 \times 3 \times \log_2{256} \  \text{bits} = 196608 \  \text{bits}$, and the leakage from the labels is comparatively negligible. For the industrial application of FL, the number of clients can be even larger (possibly up to $10^{10}$), and the privacy leakage converges to 0.

\begin{figure}
	\centering
	\includegraphics[width=0.2\linewidth]{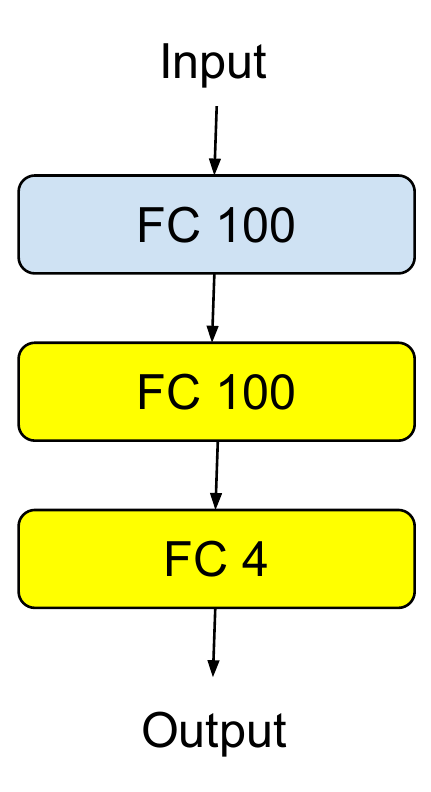}
	\caption{Diagram of the DNN model for training on dry bean dataset. The fully connected layers are denoted as "FC $\{$output size$\}$". For brevity, the ReLU and Sigmoid activation functions are not shown. }
	\label{fig:DryBeanDNN}
\end{figure}

Apart from CIFAR-10, we also show a small scale experiment on predicting the type of dry beans via 16 attributes \cite{KOKLU2020105507}. We train the source model on the data including types Seker, Horoz, Sira, and we transfer the source model to the target model which predicts the data with types Barbunya, Bombay, Cali, and Dermason. For both cases, $\frac{1}{3}$ of the dataset is randomly extracted for testing. We assume that there are $U=1176$ clients in total, and each iteration takes a protion $C=6.8 \times 10^{-3}$ of all clients, and each batch contains $K_u=4$ samples. We show the structure of considered DNN in Fig. \ref{fig:DryBeanDNN}, where we use the Adam optimizer with learning rate $7 \times 10^{-4}$. The performance of four algorithms are shown in Table \ref{tab:exp_beans}, where FbFTL achieves a four orders of magnitude (i.e., $10^{-4}$) reduction in the uplink payload compared to those of other algorithms.

\begin{table}
	\fontsize{7}{8.5}
	\selectfont
	\begin{center}
		\caption{Comparison between FL, FTL updating full model, FTL updating task-specific sub-model and FbFTL}
		\label{tab:exp_beans}
		\begin{tabular}{ | l || l | l | l | l | }
			\hline
			 & FL & FTL$_f$ & FTL$_c$ & FbFTL \\ \hline\hline
			upload batches & 41160 & 31752 & 38808 & 4703  \\ \hline
			upload parameters per batch & 12204 & 12204 & 10504 & 100  \\ \hline
			uplink payload per batch & 390.5 Kb & 390.5 Kb & 336.1 Kb & 3.2 Kb  \\ \hline
			total uplink payload $P$ & 16.1 Gb & 12.4 Gb & 15.2 Gb & 15.0 Mb  \\ \hline
			total downlink payload $D$ & 2.0 Gb & 1.6 Gb & 1.9 Gb & 336 Kb  \\ \hline
			test accuracy & 98.23\% & 98.45\% & 98.06\% & 98.06\%  \\ \hline
		\end{tabular}
	\end{center}
\end{table}

\begin{figure}
	\centering
	\includegraphics[width=0.8\linewidth]{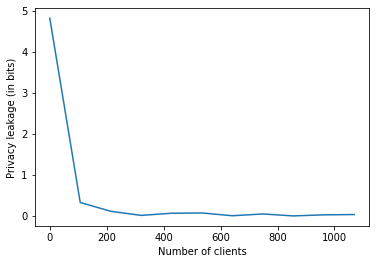}
	\caption{Privacy leakage of each batch for FbFTL in bits against the number of clients. }
	\label{fig:leakage_beans}
\end{figure}

In Fig. \ref{fig:leakage_beans}, we show the amount of privacy leakage of a certain batch in the settings with different number of clients $U$. As the number of participating clients increases, the privacy leakage approaches 0 in our considered experiment setting with 1176 clients. 

\section{Conclusion}\label{sec:con}
In this paper, we have introduced the feature-based federated transfer learning (FbFTL), as a novel communication-efficient federated learning method that uploads the features and outputs instead of gradient updates. We have described in detail the proposed system design and the learning algorithm, compared its theoretical payload with federated learning and federated transfer learning, and analyzed the privacy preservation via random shuffling. We have shown that significant reduction is achieved in both uplink and downlink payload when FbFTL is employed. Via experiments, we have further demonstrated the effectiveness of the proposed FbFTL by showing that FbFTL reduces the uplink payload by up to five orders of magnitude compared to that of existing methods. Moreover, we have  shown that with random shuffling, privacy leakage of FbFTL vanishes as the number of clients grows. These features render FbFTL a communication-efficient and privacy-preserving novel federated transfer learning scheme.

\bibliographystyle{IEEEtran}
\bibliography{ref.bib}

\end{document}